# Sensitivity Analysis in Bayesian Networks: From Single to Multiple Parameters

**Hei Chan and Adnan Darwiche**
Computer Science Department
University of California, Los Angeles
Los Angeles, CA 90095
{*hei,darwiche*}*@cs.ucla.edu*

## Abstract

Previous work on sensitivity analysis in Bayesian networks has focused on single parameters, where the goal is to understand the sensitivity of queries to single parameter changes, and to identify single parameter changes that would enforce a certain query constraint. In this paper, we expand the work to multiple parameters which may be in the CPT of a single variable, or the CPTs of multiple variables. Not only do we identify the solution space of multiple parameter changes that would be needed to enforce a query constraint, but we also show how to find the optimal solution, that is, the one which disturbs the current probability distribution the least (with respect to a specific measure of disturbance). We characterize the computational complexity of our new techniques and discuss their applications to developing and debugging Bayesian networks, and to the problem of reasoning about the value (reliability) of new information.

## 1 Introduction

Sensitivity analysis in Bayesian networks [13, 9] is broadly concerned with understanding the relationship between local network parameters and global conclusions drawn based on the network [12, 2, 8, 11]. This understanding can be useful in a number of areas, including model debugging and system design. In model debugging, the user may wish to identify parameters that are relevant to certain queries, or to identify parameter changes that would be necessary to enforce certain sanity checks on the values of probabilistic queries. In system design, sensitivity analysis can be used to choose false–positive and false–negative rates for sensors and tests to ensure the quality of an information system based on Bayesian networks.

One technical formalization of sensitivity analysis is as follows. Given a Bayesian network, and a subset of network parameters, we would like to identify possible changes to these parameters that can ensure the satisfaction of a query constraint, such as $Pr(z \mid \mathbf{e}) \geq p$, for some event $z$ and evidence $\mathbf{e}$. Other possible query constraints include $Pr(z_1 \mid \mathbf{e})/Pr(z_2 \mid \mathbf{e}) \geq k$ and $Pr(z_1 \mid \mathbf{e}) - Pr(z_2 \mid \mathbf{e}) \geq k$ for events $z_1$ and $z_2$. Figure 1 depicts an example of a sensitivity analysis session using SAMIAM [1]. Here, the network portrays an information system for predicting pregnancy based on the results of three tests. The current evidence indicates that the blood test is positive while the urine test is negative, and the probability of pregnancy given the test results is 90%. Suppose, however, that we wish the test results to confirm pregnancy to no less than 95%. Sensitivity analysis can be used in this case to identify necessary parameter changes to enforce this constraint, which can translate to changes in the false–positive and false–negative rates of various tests (which can be implemented by obtaining more reliable tests).

A key aspect of sensitivity analysis is the number of considered parameters. The simplest case involves one parameter at a time, i.e., we are only allowed to change a single parameter in the network to ensure our query constraint. Previous work has provided a procedure to find these single-parameter changes [4], using the fact that any joint probability is a linear function of any network parameter. Specifically, given a parameter $\theta_{x|\mathbf{u}}$, we can solve for the possible values of $\Delta\theta_{x|\mathbf{u}}$ that can ensure a given constraint. The time complexity needed to identify such changes for all network parameters is the same as performing inference using classical algorithms such as jointree. In our previous example, we can solve for the solution for each parameter using SAMIAM where the parameter changes are displayed in Figure 1. For example, one of the suggestions is to change the false–positive of the blood test from 10% to no more than 5%. Obviously, there are many parameters that are irrelevant to the query in this case.

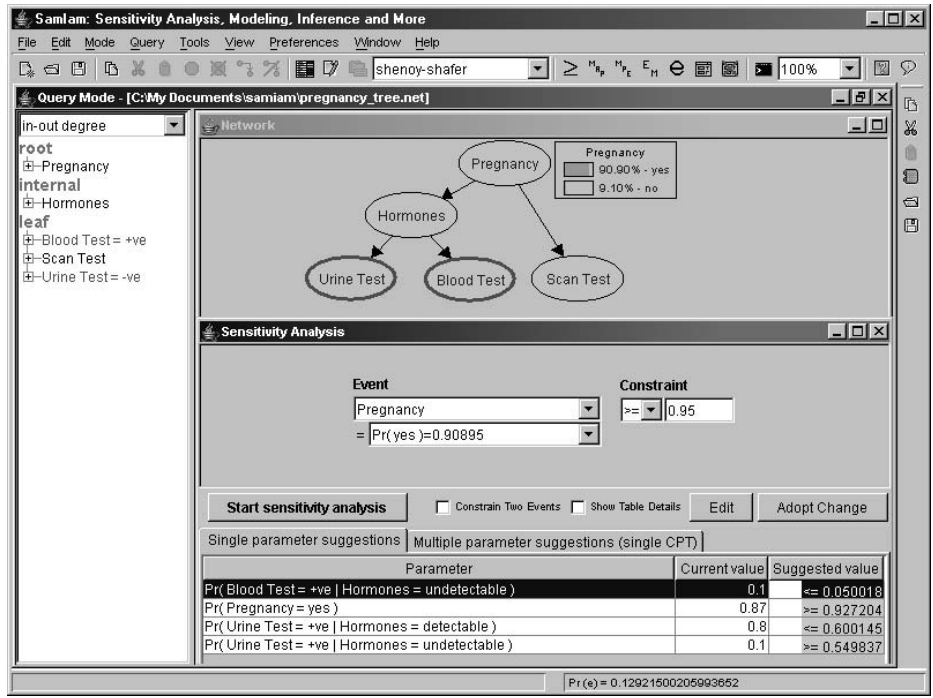

Figure 1: Finding parameter changes using the sensitivity analysis tool in SamIam.

Single parameter changes are easy to visualize and compute, but they are only a subset of possible parameter changes. We may generally be interested in changing multiple parameters in the network simultaneously to ensure the query constraint. To facilitate this, we need to understand the interaction between any joint probability and any set of network parameters [5, 6]. One common case involves changing multiple parameters but within the same conditional probability table (CPT) of some variable. The first contribution of this paper is that of showing how to identify such changes, with little extra computation beyond that needed for single–parameter changes. This is significant since multiple parameter changes can be more meaningful, and may disturb the probability distribution less significantly than single parameter changes. Practically speaking, this new technique allows us to change both the false–positive and false–negative rates of a certain information source, which can allow the enforcement of certain constraints that cannot be enforced by only changing either the false–positive or the false–negative rate.

Our second contribution involves techniques for finding parameter changes that involve multiple CPTs. However, as we will show, the complexity increases linearly in the size of each additional CPT that is involved. Therefore, practically, we can only compute suggestions of parameter changes involving a small subset of CPTs.

As expected, the solution space for multiple parameters will be a region in the $k$-dimensional space, where $k$ is the number of involved parameters. For example, for the case where we change two parameters in the same CPT, the solution space will be a half–plane, in the form of $\alpha_1 \Delta\theta_1 + \alpha_2 \Delta\theta_2 \geq c$. These results are difficult to visualize and present to users. Hence, we may want to identify and report a particular point in the solution space, i.e., a specific amount of change in $\theta_1$ and $\theta_2$. Now, the key question becomes: Which point in the solution space should we report? The approach we shall adopt is to report the point which minimizes model disturbance. But this brings another question: How to measure and quantify model disturbance?

To address this question, we will quantify the disturbance to a model by measuring the distance between the original distribution $pr$ and the new one $Pr$ (after the parameters have been changed) using a specific distance measure [3] for reasons we will discuss later.

A third contribution in this paper relates to the application of our results to the problem of reasoning about uncertain evidence. Specifically, we show how our results allow us to identify the weakest uncertain evidence, and on what network variables, that is needed to confirm a given hypothesis to some degree.

## 2 Sensitivity Analysis: Single CPT

We will present solutions to two key problems in this section. First, given a Bayesian network that specifies a distribution $pr$, and a variable $X$ with parents $\mathbf{U}$, we want to identify all changes to parameters $\theta_{x|\mathbf{u}}$ in the CPT of $X$ which would enforce the constraint $Pr(z \mid \mathbf{e}) \geq p$. Here, $Z$ and $\mathbf{E}$ are arbitrary variables in the network, $pr$ is the distribution before we apply parameter changes, and $Pr$ is the distribution after the change. Second, among the identified changes, we want to select those that minimize the distance between the old distribution $pr$ and new one $Pr$ according to the following distance measure [3]:

$$D(Pr, pr) \stackrel{def}{=} \ln \max_\omega \frac{pr(\omega)}{Pr(\omega)} - \ln \min_\omega \frac{pr(\omega)}{Pr(\omega)}. \quad (1)$$

This measure allows one to bound the amount of change in the value of any query $\beta_1 \mid \beta_2$, from $pr$ to $Pr$, as follows:

$$\frac{pr(\beta_1 \mid \beta_2)e^d}{pr(\beta_1 \mid \beta_2)(e^d - 1) + 1} \geq Pr(\beta_1 \mid \beta_2) \geq \frac{pr(\beta_1 \mid \beta_2)e^{-d}}{pr(\beta_1 \mid \beta_2)(e^{-d} - 1) + 1}, \quad (2)$$

where $d = D(Pr, pr)$. Hence, by minimizing this distance measure, we are able to provide tighter bounds on global belief changes caused by local parameter changes.

One obvious side-effect of changing the parameter $\theta_{x|\mathbf{u}}$ is that parameters $\theta_{x'|\mathbf{u}}$, for all $x' \neq x$, must also be changed such that the sum of all these parameters remain 1. Therefore, if $X$ is binary, the parameter $\theta_{\bar{x}|\mathbf{u}}$ must be changed by an equal but opposite amount. If $X$ is multi-valued, we can assume a proportional scheme of co-varying the other parameters, such that the ratio between them remain the same. However, sometimes certain parameters should remain unchanged, such as parameters who are assigned 0 values [15]. This capability is provided in SAMIAM, where users can lock certain parameters from being changed during sensitivity analysis by checking a flag. In this paper, for simplicity of presentation, we will assume that $X$ is binary with two values $x$ and $\bar{x}$, where we can obtain similar (but more wordy) results for $X$ being multi-valued [4].

### 2.1 Identifying sufficient parameter changes

We first note that the joint probability $Pr(\mathbf{e})$ can be expressed in terms of the parameters in the CPT of $X$:

$$Pr(\mathbf{e}) = C + \sum_\mathbf{u} C_\mathbf{u} \theta_{x|\mathbf{u}},$$

where $C$ is a constant, and:

$$C_\mathbf{u} = \frac{\partial Pr(\mathbf{e})}{\partial \theta_{x|\mathbf{u}}}.$$

As a reminder, $Pr(\mathbf{e})$ is linear in $\theta_{x|\mathbf{u}}$, and hence, $\partial Pr(\mathbf{e})/\partial \theta_{x|\mathbf{u}}$ is a constant, independent of the value of $\theta_{x|\mathbf{u}}$. Moreover, $\partial^2 Pr(\mathbf{e})/\partial \theta_{x|\mathbf{u}} \partial \theta_{x|\mathbf{u}'} = 0$, for any parent instantiations $\mathbf{u}$ and $\mathbf{u}'$ [6]. Therefore, if we apply a change of $\Delta\theta_{x|\mathbf{u}}$ to each $\theta_{x|\mathbf{u}}$, we have:

$$\begin{aligned}
\Delta Pr(\mathbf{e}) &= Pr(\mathbf{e}) - pr(\mathbf{e}) \\
&= \sum_\mathbf{u} C_\mathbf{u} \Delta \theta_{x|\mathbf{u}} \\
&= \sum_\mathbf{u} \frac{\partial Pr(\mathbf{e})}{\partial \theta_{x|\mathbf{u}}} \Delta \theta_{x|\mathbf{u}}. \quad (3)
\end{aligned}$$

Now, to find the solution of parameter changes that satisfies $Pr(z \mid \mathbf{e}) \geq p$ (or equivalently, $Pr(z, \mathbf{e}) \geq p \cdot Pr(\mathbf{e})$), the following must hold:

$$\Delta Pr(z, \mathbf{e}) + pr(z, \mathbf{e}) \geq p(\Delta Pr(\mathbf{e}) + pr(\mathbf{e})).$$

From Equation 3, we have:

$$\begin{aligned}
&\sum_\mathbf{u} \frac{\partial Pr(z, \mathbf{e})}{\partial \theta_{x|\mathbf{u}}} \Delta \theta_{x|\mathbf{u}} + pr(z, \mathbf{e}) \\
&\geq p \left( \sum_\mathbf{u} \frac{\partial Pr(\mathbf{e})}{\partial \theta_{x|\mathbf{u}}} \Delta \theta_{x|\mathbf{u}} + pr(\mathbf{e}) \right).
\end{aligned}$$

Rearranging the terms, we get:

$$\sum_\mathbf{u} \alpha(\theta_{x|\mathbf{u}}) \Delta \theta_{x|\mathbf{u}} \geq -(pr(z, \mathbf{e}) - p \cdot pr(\mathbf{e})), \quad (4)$$

where:

$$\alpha(\theta_{x|\mathbf{u}}) = \frac{\partial Pr(z, \mathbf{e})}{\partial \theta_{x|\mathbf{u}}} - p \frac{\partial Pr(\mathbf{e})}{\partial \theta_{x|\mathbf{u}}}. \quad (5)$$

The first problem addressed in this section can then be solved by finding possible combinations of $\Delta \theta_{x|\mathbf{u}}$ that satisfy Inequality 4. The solution space can be found by solving for the equality condition, and it will be in the shape of a half–space due to the linearity of our terms.

To find the solution space of Inequality 4, we need to compute all partial derivatives of the form $\partial Pr(z, \mathbf{e})/\partial \theta_{x|\mathbf{u}}$ and $\partial Pr(\mathbf{e})/\partial \theta_{x|\mathbf{u}}$. They can be computed using the jointree algorithm [14, 10] or the differential approach [6]. The time complexity of this computation is $O(n \exp(w))$, where $w$ is the network treewidth, and $n$ is the number of network parameters [6]. This complexity is the same as that of computing the probability of evidence $Pr(\mathbf{e})$. Moreover, the above method generalizes a previous method [4] for identifying single parameter changes, yet has the same complexity!

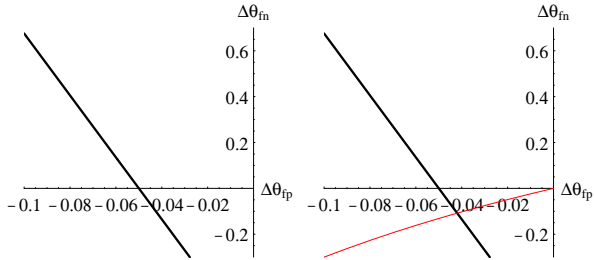

Figure 2: Finding single CPT changes for Example 2.1. On the left, we plot the solution space in terms of $\Delta\theta_{fp}$ and $\Delta\theta_{fn}$, which is the region below the line. On the right, we illustrate how we find the optimal solution, by moving on the curve where the log-odds change in the two parameters are the same, as the optimal solution is the intersection of the line and the curve.

**Example 2.1** *Given the sensitivity analysis problem shown in Figure 1, we are now interested in changing multiple parameters in a single CPT to satisfy the constraint. For example, we may want to use a more reliable blood test to satisfy our desired constraint. Currently, the false–positive of the test is* 10%, *while the false–negative is* 30%. *We will denote these two parameters as $\theta_{fp}$ and $\theta_{fn}$ respectively. We can find the $\alpha$ terms for both parameters given by Equation 5, and plug into Inequality 4:*

$$-.1061\Delta\theta_{fp} - .0076\Delta\theta_{fn} \geq .0053.$$

*The solution space is plotted on the left of Figure 2. The line indicates the set of points where the equality condition holds, while the solution space is the region below the line. Therefore, any parameter changes in this region will be able to ensure the constraint that the probability of pregnancy given the test results is at least* 95%.

### 2.2 Identifying optimal parameter changes

We now address the second problem of interest in this section: identifying the solution of Inequality 4 that minimizes the distance between the original and new distribution. This solution is unique and can be identified using a simple local search procedure. But the technique is based on several observations.

First, since the new distribution $Pr$ is obtained from $pr$ by changing only one CPT, the distance between $pr$ and $Pr$ is exactly the distance between the old and new CPT (each viewed as a distribution) [3].

Second, there is a closed form for this distance:[1]

$$D_{X|\mathbf{U}} = \max_{\mathbf{u}} \left| \log \frac{\theta_{x|\mathbf{u}} + \Delta\theta_{x|\mathbf{u}}}{1 - (\theta_{x|\mathbf{u}} + \Delta\theta_{x|\mathbf{u}})} - \log \frac{\theta_{x|\mathbf{u}}}{1 - \theta_{x|\mathbf{u}}} \right|. \quad (6)$$

Note that the quantity being maximized above is nothing but the absolute log–odds change for parameter $\theta_{x|\mathbf{u}}$, i.e., $|\Delta(\log O(\theta_{x|\mathbf{u}}))|$.

Third, we must be able to find an optimal solution on the line where $Pr(z \mid \mathbf{e}) = p$, since if there is an optimal solution where $Pr(z \mid \mathbf{e}) > p$, we can always decrease the absolute log–odds change in some parameter to satisfy the equality condition, and the distance measure will not increase.

Finally, for any solution that satisfies $Pr(z \mid \mathbf{e}) = p$, it follows from Equation 6 that the solution which minimizes the distance $D_{X|\mathbf{U}}$ is the one where the absolute log–odds changes in all parameters in the CPT is the same. This is because to obtain another solution on the line, we must increase the absolute log–odds change in one parameter and decrease it in another, thereby producing a larger distance measure.

Given the above observations, we can now search for the optimal single CPT parameter change that satisfies Inequality 4 using the following local search procedure:

1. Pick all parameters $\theta_{x|\mathbf{u}}$ in the CPT of $X$ where the terms $\alpha(\theta_{x|\mathbf{u}})$ are non–zero, and categorize them according to whether the term is positive or negative.

2. Pick a certain amount of absolute log–odds change $|\Delta(\log O(\theta_{x|\mathbf{u}}))|$, and apply it to each parameter. Whether a parameter is increased or decreased depends on its $\alpha$ term.

3. If $Pr(z \mid \mathbf{e}) = p$ within an acceptable degree of error, we have found the optimal solution. Otherwise, try a larger $|\Delta(\log O(\theta_{x|\mathbf{u}}))|$ if $Pr(z \mid \mathbf{e}) < p$, or a smaller $|\Delta(\log O(\theta_{x|\mathbf{u}}))|$ if $Pr(z \mid \mathbf{e}) > p$. The new amount of $|\Delta(\log O(\theta_{x|\mathbf{u}}))|$ applied should be determined numerically by the new query value of $Pr(z \mid \mathbf{e})$ for a fast rate of convergence.

On the right of Figure 2, we provide an illustration of our procedure applied to Example 2.1. There are two parameters in the CPT we are allowed to change, the false–positive and the false–negative rates. The region below the line is the solution space. The points on the new curve are those where the log–odds changes in the two parameters are the same (both parameters are decreased because their $\alpha$ terms are negative). To find the optimal solution, we only need to move on this curve using a numerical method, until we are at the intersection of the line and the curve. It is given by $\Delta\theta_{fp} = -.042$ and $\Delta\theta_{fn} = -.109$, i.e., the new false–positive should be 5.8% and the new false–negative should be 19.1%.

---
[1]Equation 6 assumes $Pr(\mathbf{u}) > 0$ for all instantiations $\mathbf{u}$. If $Pr(\mathbf{u}) = 0$ for some $\mathbf{u}$, any query value will not respond to any change in the parameter $\theta_{x|\mathbf{u}}$, so we can leave it out when computing the distance measure.

The above technique has been implemented in SAMIAM which is available for download [1].

## 3 Single Parameter Changes vs. Single CPT Changes

In this section, we make comparisons between single parameter changes and single CPT changes. As we have shown, both types of suggestions require the same amount of computations to find, in computing the partial derivatives of joint probabilities with respect to all parameters. However, solutions of single CPT changes are harder to visualize and present, and it takes a little more time to find the optimal solution using the numerical method we proposed.

However, it is advantageous to apply single CPT changes instead of single parameter changes to a Bayesian network in order to satisfy a query constraint. First, single CPT changes are more meaningful and intuitive than single parameter changes. For example, given a sensor in a network, single parameter changes amount to changing only the false–positive or false–negative rate of this sensor, while single CPT changes allow one to change both rates.

Second, for some variable in the network, there may exist single CPT changes, but not single parameter changes, that can ensure a certain query constraint. For an example consider Figure 4, which depicts a Bayesian network involving a scenario of potential fire in a building. Currently, we are given evidence of smoke and people leaving the building, and the probability that there is a tampering of the alarm given the evidence is 2.87%. We may now pose the question: what parameter changes can we apply to decrease this value to at most 1%?

If we can only change a single parameter in the network, SAMIAM returns a simple answer: the only parameter you can change is the prior probability of tampering, from 2% to .7%. You cannot change any single parameter in the CPT of the Alarm variable (representing whether the alarm is triggered, by fire, tampering or other sources) to ensure the constraint, and we may be inclined to believe that the parameters in this CPT are irrelevant to the query. However, if we are allowed to change multiple parameters in a single CPT, SAMIAM returns a new suggestion, telling us that we can indeed change the CPT of the Alarm variable to ensure our constraint. The optimal suggestion computed by SAMIAM is shown in Figure 4, where the original parameter values are in white background, and the suggested parameter values are in shaded background. The distance measure of this parameter change is 2.29.

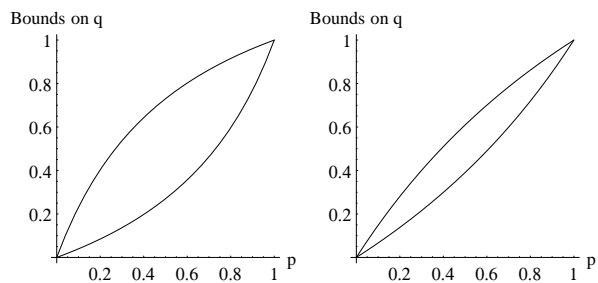

Figure 3: The plot of the bounds on the new value of any query, $q = Pr(\beta_1 \mid \beta_2)$, in terms of its original value $p = pr(\beta_1 \mid \beta_2)$, for the suggested single parameter change, with $d = .995$, and the suggested single CPT change, with $d = .445$.

Finally, even if changes of both types are available, single CPT changes are often preferred because they disturb the network less significantly, as they incur a smaller distance measure. For example, we can pose another query constraint, where we want to decrease the query value from 2.87% to at most 2.5%. This time, for the CPT of the Alarm variable, SAMIAM returns parameter change suggestions of both types. A possible single parameter change is to decrease the probability of the alarm triggered given tampering but no fire from 85% to 67.7%, incurring a distance measure of .995. On the other hand, if we change all parameters in the CPT simultaneously, the distance measure incurred is a much smaller value of .445.

From Inequality 2, the distance measure computed for a parameter change quantifies the disturbance to the original probability distribution, by providing bounds on changes in any query $\beta_1 \mid \beta_2$. In Figure 3, we plot the bounds on the new value of any query, $q = Pr(\beta_1 \mid \beta_2)$, in terms of its original value, $p = pr(\beta_1 \mid \beta_2)$, for the respective values of the distance incurred by the suggested single parameter change and the suggested single CPT change respectively. As we can see, the suggested single CPT change ensures a tighter bound on the change in any query value.

## 4 Sensitivity Analysis: Multiple CPTs

In this section, we allow the changing of parameters in multiple CPTs simultaneously. For example, we may want to change all parameters in the CPTs of variables $X$ and $Y$, whose parents are $\mathbf{U}$ and $\mathbf{V}$ respectively. In this case, the joint probability $Pr(\mathbf{e})$ can be expressed in terms of the parameters in both CPTs as:

$$Pr(\mathbf{e}) = C + \sum_{\mathbf{u}} C_{\mathbf{u}} \theta_{x|\mathbf{u}} + \sum_{\mathbf{v}} C_{\mathbf{v}} \theta_{y|\mathbf{v}} + \sum_{\mathbf{u},\mathbf{v}} C_{\mathbf{u},\mathbf{v}} \theta_{x|\mathbf{u}} \theta_{y|\mathbf{v}},$$

where $C$ is a constant, and:

$$\frac{\partial Pr(\mathbf{e})}{\partial \theta_{x|\mathbf{u}}} = C_{\mathbf{u}} + \sum_{\mathbf{v}} C_{\mathbf{u},\mathbf{v}} \theta_{y|\mathbf{v}};$$

$$\frac{\partial Pr(\mathbf{e})}{\partial \theta_{y|\mathbf{v}}} = C_{\mathbf{v}} + \sum_{\mathbf{u}} C_{\mathbf{u},\mathbf{v}} \theta_{x|\mathbf{u}};$$

$$\frac{\partial^2 Pr(\mathbf{e})}{\partial \theta_{x|\mathbf{u}} \partial \theta_{y|\mathbf{v}}} = C_{\mathbf{u},\mathbf{v}}.$$

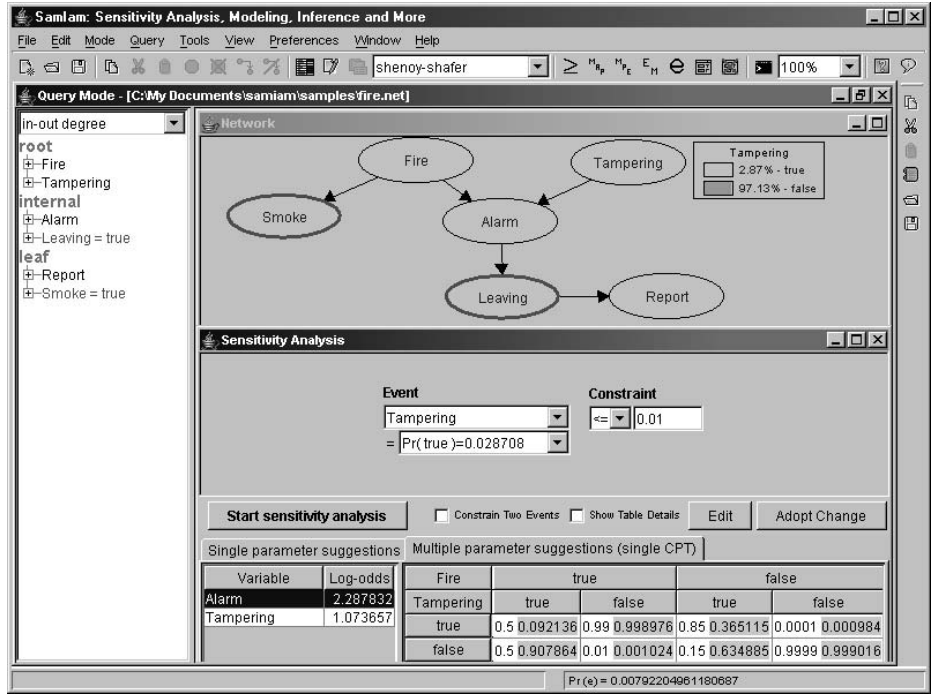

Figure 4: Finding single CPT changes using the sensitivity analysis tool of SAMIAM.

Therefore, if we apply a change of $\Delta\theta_{x|\mathbf{u}}$ to each $\theta_{x|\mathbf{u}}$, and a change of $\Delta\theta_{y|\mathbf{v}}$ to each $\theta_{y|\mathbf{v}}$, the change in the joint probability $Pr(\mathbf{e})$ is given by:

$$\Delta Pr(\mathbf{e}) = \sum_{\mathbf{u}} \left( C_{\mathbf{u}} + \sum_{\mathbf{v}} C_{\mathbf{u},\mathbf{v}} \theta_{y|\mathbf{v}} \right) \Delta\theta_{x|\mathbf{u}}$$
$$+ \sum_{\mathbf{v}} \left( C_{\mathbf{v}} + \sum_{\mathbf{u}} C_{\mathbf{u},\mathbf{v}} \theta_{x|\mathbf{u}} \right) \Delta\theta_{y|\mathbf{v}}$$
$$+ \sum_{\mathbf{u},\mathbf{v}} C_{\mathbf{u},\mathbf{v}} \Delta\theta_{x|\mathbf{u}} \Delta\theta_{y|\mathbf{v}}.$$

$$= \sum_{\mathbf{u}} \frac{\partial Pr(\mathbf{e})}{\partial \theta_{x|\mathbf{u}}} \Delta\theta_{x|\mathbf{u}} + \sum_{\mathbf{v}} \frac{\partial Pr(\mathbf{e})}{\partial \theta_{y|\mathbf{v}}} \Delta\theta_{y|\mathbf{v}}$$
$$+ \sum_{\mathbf{u},\mathbf{v}} \frac{\partial^2 Pr(\mathbf{e})}{\partial \theta_{x|\mathbf{u}} \partial \theta_{y|\mathbf{v}}} \Delta\theta_{x|\mathbf{u}} \Delta\theta_{y|\mathbf{v}}. \quad (7)$$

Now, to find the solution of parameter changes that satisfies $Pr(z \mid \mathbf{e}) \geq p$, from Equation 7, we have:

$$\sum_{\mathbf{u}} \frac{\partial Pr(z,\mathbf{e})}{\partial \theta_{x|\mathbf{u}}} \Delta\theta_{x|\mathbf{u}} + \sum_{\mathbf{v}} \frac{\partial Pr(z,\mathbf{e})}{\partial \theta_{y|\mathbf{v}}} \Delta\theta_{y|\mathbf{v}}$$
$$+ \sum_{\mathbf{u},\mathbf{v}} \frac{\partial^2 Pr(z,\mathbf{e})}{\partial \theta_{x|\mathbf{u}} \partial \theta_{y|\mathbf{v}}} \Delta\theta_{x|\mathbf{u}} \Delta\theta_{y|\mathbf{v}} + pr(z,\mathbf{e})$$
$$\geq p \left( \sum_{\mathbf{u}} \frac{\partial Pr(\mathbf{e})}{\partial \theta_{x|\mathbf{u}}} \Delta\theta_{x|\mathbf{u}} + \sum_{\mathbf{v}} \frac{\partial Pr(\mathbf{e})}{\partial \theta_{y|\mathbf{v}}} \Delta\theta_{y|\mathbf{v}} \right.$$
$$\left. + \sum_{\mathbf{u},\mathbf{v}} \frac{\partial^2 Pr(\mathbf{e})}{\partial \theta_{x|\mathbf{u}} \partial \theta_{y|\mathbf{v}}} \Delta\theta_{x|\mathbf{u}} \Delta\theta_{y|\mathbf{v}} + pr(\mathbf{e}) \right).$$

Rearranging terms, we get:

$$\sum_{\mathbf{u}} \alpha(\theta_{x|\mathbf{u}}) \Delta\theta_{x|\mathbf{u}} + \sum_{\mathbf{v}} \alpha(\theta_{y|\mathbf{v}}) \Delta\theta_{y|\mathbf{v}}$$
$$+ \sum_{\mathbf{u},\mathbf{v}} \alpha(\theta_{x|\mathbf{u}}, \theta_{y|\mathbf{v}}) \Delta\theta_{x|\mathbf{u}} \Delta\theta_{y|\mathbf{v}}$$
$$\geq -(pr(z,\mathbf{e}) - p \cdot pr(\mathbf{e})), \quad (8)$$

where $\alpha(\theta_{x|\mathbf{u}})$ and $\alpha(\theta_{y|\mathbf{v}})$ are given by Equation 5, and:

$$\alpha(\theta_{x|\mathbf{u}}, \theta_{y|\mathbf{v}}) = \frac{\partial^2 Pr(z, \mathbf{e})}{\partial \theta_{x|\mathbf{u}} \partial \theta_{y|\mathbf{v}}} - p \frac{\partial^2 Pr(\mathbf{e})}{\partial \theta_{x|\mathbf{u}} \partial \theta_{y|\mathbf{v}}}. \quad (9)$$

Therefore, additionally we need to compute the second partial derivatives of $Pr(z, \mathbf{e})$ and $Pr(\mathbf{e})$ with respect to $\theta_{x|\mathbf{u}}$ and $\theta_{y|\mathbf{v}}$ for all pairs of $\mathbf{u}$ and $\mathbf{v}$. A simple way to do this would be to set evidence on every family instantiation $x, \mathbf{u}$, then find the derivatives with respect to $\theta_{y|\mathbf{v}}$ for all $\mathbf{v}$ [6]. The complexity of this method is $O(nF(X)\exp(w))$, where $F(X)$ is the number of family instantiations of $X$, i.e., the size of the CPT. This approach is however limited to non–extreme values of $\theta_{x|\mathbf{u}}$, yet it allows one to use any general inference algorithm [6]. For extreme parameters, one can use a specific inference approach [6] to obtain these derivatives using the same complexity as given above.

The computations above can be expanded to multiple parameter changes involving more than two CPTs. For example, if we change three CPTs simultaneously, we need to compute the third partial derivatives with respect to the corresponding parameters. The complexity of obtaining these higher order derivatives is $O(n \prod_{X_i} F(X_i) \exp(w))$, where $X_i$ are the variables whose CPTs we are interested in [6].

**Example 4.1** *We again refer to the fire network, and pose another sensitivity analysis problem. Given evidence that people are leaving but no smoke is observed, the current probability of having a fire is* 5.2%. *We wish to constrain this query value to at most* 2.5%. SAMIAM *indicates that we can accomplish this by decreasing the prior probability of fire, $\theta_F$, from* 1% *to* .47%, *or increasing the prior probability of tampering, $\theta_T$, from* 2% *to* 4.39%. *However, what are the changes necessary if we are allowed to change both parameters?*

*To answer this, we find the $\alpha$ terms given by Equations 5 and 9, and plug into Inequality 8:*

$$-.0845\Delta\theta_F + .0187\Delta\theta_T - .7816\Delta\theta_F\Delta\theta_T \geq .000448.$$

*The solution space is plotted on the left of Figure 5. The curve indicates the set of points where the equality condition holds, while the solution space is the region above the curve. Therefore, any parameter changes in this region will be able to ensure the constraint that the probability of fire given the evidence is at most* 2.5%.

We now wish to compute the distance measure for parameter change suggestions involving multiple CPTs, in order to find the optimal solution. Although this cannot be easily computed in some cases, for the cases where the families $X, \mathbf{U}$ and $Y, \mathbf{V}$ are disjoint, i.e., $X$ and $Y$ do not have a parent-child relationship and do

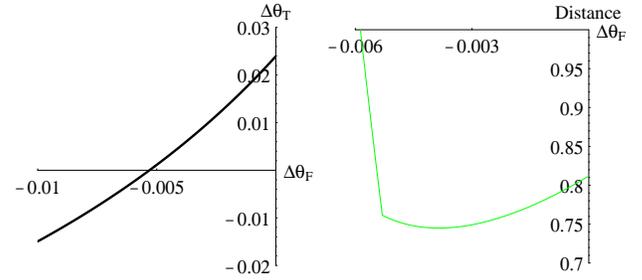

Figure 5: Finding multiple CPT changes for Example 4.1. On the left, we plot the solution space in terms of $\Delta\theta_F$ and $\Delta\theta_T$, which is the region below the curve. On the right, we illustrate how we find the optimal solution, by computing the distance measure for each point on the curve in terms of $\theta_F$, and locating the minimum.

not have a common parent, the distance measure can be easily computed as [3]:

$$D_{\{X|\mathbf{U}, Y|\mathbf{V}\}} = D_{X|\mathbf{U}} + D_{Y|\mathbf{V}}. \quad (10)$$

Here, the total distance measure can be computed as the sum of the distances caused individually by each of the CPT changes, as computed by Equation 6.[2] Even though we have this restriction of disjointness for Equation 10, many CPTs satisfy this condition. For example, the two variables, Fire and Tampering, involved in Example 4.1, are both roots, and hence, satisfy our condition. Moreover, when the variables involved are sensors on different variables in a Bayesian network, their families are disjoint, and we can easily compute the distance measure using Equation 10.

Similar to single CPT changes, we are often more interested in finding the optimal solution than presenting the whole solution space. As in the previous case, we can find an optimal solution on the curve where $Pr(z \mid \mathbf{e}) = p$, and also with the property that the log-odds changes in the parameters of each individual CPT are the same. With these two assumptions, we can find the combination of CPT changes that gives us the smallest distance measure.

For example, we can find the optimal solution for Example 4.1 by traversing on the curve where $Pr(z \mid \mathbf{e}) = p$, and searching for the point with the smallest distance measure. On the right of Figure 5, we plot the distance measure for points on that curve in terms of $\Delta\theta_F$. The minimum is attained when $\Delta\theta_F = -.0039$ and $\Delta\theta_T = .0056$, i.e., the new prior probabilities are .61% and 2.56% respectively. The distance measure given by the optimal solution is .745.

---
[2]If the families $X, \mathbf{U}$ and $Y, \mathbf{V}$ are not disjoint, the distance measure cannot be computed as the individual sums, because a pair of instantiations of the two CPTs may not be consistent. In this case, we can still compute the distance measure using a procedure which multiplies two tables (thereby eliminating inconsistent pairs of instantiations), a harder but still manageable process.

Because of the computations involved in finding solutions involving multiple CPTs, the key to any automated sensitivity analysis tool which implements this procedure is to find relevant CPTs to check for solutions, instead of trying all combinations of CPTs, which would be computationally too costly. The first partial derivatives computed for finding single CPT changes can serve as a guide for identifying these relevant CPTs. For many CPTs, the first partial derivatives with respect to the parameters are 0, eliminating them from consideration. On the other hand, we should definitely consider CPTs where small parameters changes can induce large changes in the user-selected queries. Hence, the numerical procedure is not as straightforward as the one for single CPT changes.

## 5 Searching for the Optimal Soft Evidence

One application of sensitivity analysis with multiple parameters is to find the optimal soft evidence that would enforce a certain constraint. Soft evidence is formally defined as follows. Given two events of interest, $q$ (virtual event) and $r$ (hypothesis), we specify the soft evidence that $q$ bears on $r$ by the likelihood ratio, $\lambda = Pr(q \mid r)/Pr(q \mid \bar{r})$ [13, 7]. The virtual event $q$ serves as soft evidence on $r$, representing a partial confirmation or denial of $r$. If $\lambda$ is more than 1, $q$ argues for $r$, while if $\lambda$ is less than 1, $q$ argues against $r$. If $\lambda$ equals 1, $q$ is trivial and does not shed any new information on $r$. The likelihood ratio $\lambda$ also quantifies the strength of the soft evidence, with values closer to infinity or zero indicating more convincing arguments.

From a Bayesian network perspective, the virtual evidence $q$ can be implemented as a dummy node $Q$ which is added as a child of the variable $R$ it is reporting on. The likelihood ratio $\lambda$ will be encoded in the CPT of $Q$ by specifying $\lambda = \theta_{q|r}/\theta_{q|\bar{r}}$. The soft evidence is then incorporated by setting the value of $Q$ to $q$ [13].

For example, in the fire network we showed previously, we can add a smoke detector to the network, which generates a sound when it detects smoke, but is not perfect and is associated with small false–positive and false–negative rates. The triggering of the detector can be viewed as soft evidence on the presence of smoke, as it argues for the presence of smoke.

Given a number of variables that we can potentially gather soft evidence on, we may be interested in finding the minimum amount of soft evidence to ensure a certain query constraint. To do that, we must first add a child $Q_i$ to each variable $R_i$ of interest, set the CPT of each $Q_i$ such that all parameters are trivial, i.e., 50%, and then observe the evidence $q_i$ for every $Q_i$. Doing this will not have any impact on the results of any queries. We then run the sensitivity analysis procedure on multiple parameters, and find the optimal solution of parameter changes, restricted to parameters in the CPTs of variables $Q_i$. This solution gives us the optimal combination of soft evidence on variables $R_i$, as it minimizes the distance measure, and hence, disturbs the network least significantly.

For an example, we go back to the fire network, where we now face a scenario that the alarm is triggered. The probability of having a fire is now 36.67%. We now wish to install a smoke detector, such that when it is also triggered, the probability of fire is at least 80%. To find the reliability required for this detector, we add a "detector" node as a child of the Smoke variable, while setting all its parameters as trivial. We then add the observation of the detector being triggered as part of evidence, and perform our sensitivity analysis procedure. The result suggests that if the false–positive and the false–negative rates of the detector are both 10.98%, the reliability of the hypothesis is achieved. This is equivalent to having soft evidence on the presence of smoke with a likelihood ratio of $\lambda = 8.113$.

## 6 Conclusion

This paper made contributions to the problem of sensitivity analysis in Bayesian networks with respect to multiple parameter changes. Specifically, we presented the technical and practical details involved in identifying multiple parameter changes that are needed to satisfy query constraints. The main highlight was the ability to identify optimal, multiple parameter changes that are restricted to a single CPT, where we showed the complexity of achieving this is similar to the one for single parameter changes, except for an additional cost involved with a simple numerical method. We also addressed the problem when multiple CPTs are involved, where we characterized the corresponding solution and its (higher) complexity. Finally, we discussed a number of applications of these results, including model debugging and information–system design.

## Acknowledgments

This work has been partially supported by NSF grant IIS-9988543 and MURI grant N00014-00-1-0617.